\documentclass[]{ceurart}

\usepackage{listings}
\usepackage{amsmath,amssymb}
\usepackage{booktabs}
\usepackage{tikz}
\usetikzlibrary{positioning, arrows.meta}

\begin{document}

\copyrightyear{2026}
\copyrightclause{Copyright for this paper by its authors.
  Use permitted under Creative Commons License Attribution 4.0
  International (CC BY 4.0).}

\conference{ITAT 2026: Information Technologies -- Applications and Theory,
  September 25--29, 2026, Vr\v{s}atec, Slovakia}

\title{AutoGraphForge: Towards Automated Graph Theory Discovery}

\author[1]{J\'an Pastorek}[%
email=jan.pastorek@fmph.uniba.sk,
orcid=0000-0001-8237-1275,
]
% \cormark[1]
% \author[1]{Samuel Varchol}[%
% email=varchol3@uniba.sk,
% ]
% \author[1]{Vladim\'ir Jan\v{c}\'ar}[%
% email=jancar12@uniba.sk,
% ]
\address[1]{Department of Applied Informatics, Comenius University in
  Bratislava, Bratislava, Slovakia}

% \cortext[1]{Corresponding author.}

\begin{abstract}
  We report on our ongoing project to develop a computational pipeline \emph{AutoGraphForge} for an automated
  graph-theoretic conjecturing-refuting-formalizing-proving system. Generation of conjectures is counterexample-guided and runs in rounds: a \textsc{Graffiti3} generator
proposes conjectures over a small, evolving snapshot table $T$ (initially a few hundred
graphs with their computed invariants) that grows only by counterexamples to its own conjectures.  A \emph{novelty filter} of $559$ classical and folklore relations, closed under transitive composition and linear identity substitution, decides via a linear program whether a candidate conjecture is already implied
by known results. Then the conjecture is tested against a dataset of $\sim\!348{,}000$
  graphs unioning the complete House of Graphs invariant export, the exhaustive census of all connected
  graphs on at most nine vertices, and several extremal graph families
  (strongly regular, minimal Ramsey, Cayley, cages, barbells, lollipops, spiders, etc.), augmented by graphs generated by random models (random-regular, $G(n,p)$, random trees and random bipartite graphs). Then, \emph{counterexample-search algorithms} attempt to find counterexamples to surviving conjectures. Run for several rounds on an HPC
cluster, the loop yields ${6{,}522}$ surviving conjectures that no graph
in the refutation dataset, which were not flagged as known by the novelty filter, and no active-search run
eliminated---among them nontrivial relations between the annihilation number and the edge-cover number for bipartite graphs and regular graphs which we were able to prove by hand. A subsequent \emph{formalization and proving} stage deterministically translates each surviving conjecture into a Lean~4 statement skeleton; every candidate proof is kernel-verified against a pinned \texttt{mathlib4} and our custom invariant preamble. The formalization-and-proving stage integrates two state-of-the-art neural provers---\textsc{DeepSeek-Prover-V2-671B} (served with vLLM) and the Lean-specialised \textsc{OProver-32B}---behind this independent kernel check. This stage is implemented end-to-end and passes initial sanity checks---the deterministic export yields type-correct Lean statements and the provers independently kernel-verify trivial inequalities---with the full pipeline currently running on the cluster.
\end{abstract}


\maketitle

\section{Introduction}

Computer-assisted discovery of mathematical results and conjectures has a long history which is well documented in \cite{larson2017automated,delavina2005history}. With the advent of artificial intelligence and high-performance computing, it might be expected that the role of computers in mathematical discovery will increase. In fact, T. Tao \cite{tao2025machine} notes that new methods for computer assistance in mathematical research already include using machine learning to discover new relationships, searching for counterexamples, and using formal proof assistants to verify proofs. Moreover, as the same author notes, large language models such as ChatGPT can be used to assist in mathematical research, for example by searching through the known literature or suggesting proof strategies. A quite thorough entry point into this growing area is the proceedings of the June~2023 National Academies workshop on ``AI to Assist Mathematical Reasoning''~\cite{koretsky2023ai}; as one of its outcomes, Talia Ringer led the compilation of a community resource list on AI for mathematics, available and updated at \url{https://docs.google.com/document/d/1kD7H4E28656ua8jOGZ934nbH2HcBLyxcRgFDduH5iQ0}.

For the purpose of this paper, we divide computer-assistance in mathematical discovery based on the research stage computers assist: (i) \emph{conjecturing} new statements based on a knowledge base, (ii) \emph{refuting} them by searching for counterexamples or inconsistencies, (iii) \emph{formalizing} them into a rigorous language, and ultimately (iv) \emph{proving} them in a formal system.

Automated conjecturing in graph theory has a forty-year history, from
Fajtlowicz's \textsc{Graffiti}~\cite{fajtlowicz1988conjectures,wiki2026graffiti,larson2002intelligent} and
DeLaVi\~na's \textsc{Graffiti.pc}~\cite{delavina2002graffitipc,delavina2005history}
to the \textsc{AutoGraphiX}/\textsc{GraPHedron} programme of Aouchiche,
Caporossi, Hansen, and M\'elot~\cite{aouchiche2009variable,caporossi2004graphedron,melot2008facet} and,
more recently, Davila's \textsc{TxGraffiti}~\cite{davila2024automated,caro2022txgraffiti}
and \textsc{The Optimist}~\cite{davila2024optimist}, Larson's Dalmatian-based
\textsc{Conjecturing}~\cite{larson2016dalmatian,larson2017automated} and
\textsc{Graffiti}$^3$~\cite{davila2026graffiti3}, and the polyhedral
\textsc{PHOEG} system~\cite{bonte2026phoeg}. These systems propose inequalities
between graph invariants---$\alpha(G)\le\nu(G)$, $\chi(G)\le\Delta(G)+1$, and
so on---by fitting candidate bounds to a finite dataset of graphs and retaining and sorting only those that pass a set of filters and heuristics. Three distinct paradigms have emerged for searching the space of graph-theoretic
inequalities. \emph{Algebraic expression trees}, used by \textsc{Graffiti} and \textsc{Conjecturing}, build
candidate conjectures as rooted trees of invariants combined by unary and binary
operations (sums, products, square roots), and enumerate candidates by
exhaustive or heuristic tree search. \emph{Linear and mixed-integer programming},
used by \textsc{TxGraffiti}~\cite{davila2024automated,caro2022txgraffiti},
\textsc{The Optimist}~\cite{davila2024optimist}, and
\textsc{Graffiti}$^3$~\cite{davila2026graffiti3}, instead solves an
optimisation model over a precomputed tabular dataset: by minimising the
distance between a target invariant and a linear combination of others, the
tightest possible bound is found without enumeration. The
\emph{polyhedral/geometric} approach of \textsc{GraPHedron}~\cite{melot2008facet,caporossi2004graphedron}, 
\textsc{PHOEG}~\cite{bonte2026phoeg,christophe2008linear} and recently also \textsc{Graffiti}$^3$~\cite{davila2026graffiti3} embeds each graph as a
point in invariant space and reads off conjectures as facets of the resulting
convex hull; this produces the complete set of optimal linear
inequalities for those invariants and makes optimality a geometric certificate. Many resulting conjectures 
have become theorems; some have been refuted by hand or by machine
search, see for instance~\cite{brewster1995computational,larson2017automated,jooken2025computer}.

Anyone who uses such a system quickly encounters five possible failure modes.
First, a candidate can hold on every graph in a finite dataset yet be
\emph{false in general}: the dataset simply lacks the structure that breaks
it. Second, the system can produce a candidate that is \emph{true but known}, already recorded in the literature. Third, the system can produce a candidate that is \emph{true but trivial}, resulting in an ``avalanche of triviality''. Fourth is the \emph{monster conjecture} \cite{davila2026graffiti3}: a single
conjecture that overfits the table by accumulating many terms and constants
(producing unnecessarily complex bounds such as $\alpha\le 0.31\,\Delta+0.42\,\nu-0.07\,m+1.8$).
Fifth, the system can ``discover'' a counterexample
to a well-established conjecture, which almost always signals an error in an
invariant routine. The first and the last failure
modes are about trust: a generated inequality that ``holds'' and a
verified ``counterexample'' are each only as reliable as the data and invariant computations behind them. The second and the third failure modes are about novelty: a candidate that is true but known, or true but trivial, is mathematically uninteresting and of little value to the practitioner. All of these failure modes are mitigated---though, as \S\ref{sec:worked} shows for triviality, not yet eliminated---in our pipeline by the adversarial counterexample search, the novelty filter, and the selection heuristics.

To address the first and fourth failure modes---false bounds and overfitting---a robust refutation engine is required. Computational methods used for searching for (counter)examples of graphs have been recently summarised in a review article of Jorik Jooken~\cite{jooken2025computer}. Among others, the methods include exhaustive search with branch and bound, randomised search, and heuristic search. More recently, deep reinforcement learning has been applied to refute several conjectures \cite{wagner2021constructions}.

As for the last two stages. A \emph{proof assistant}
such as Lean~4~\cite{moura2015lean} is a programming language in which mathematical statements and
their proofs are written as formal objects; a small, trusted \emph{kernel} then mechanically checks
each proof, so a statement is accepted only if its proof is correct down to the axioms---there is no
room for a hand-waved step. Lean is paired with \texttt{mathlib4}~\cite{vandoorn2020maintaining}, a
large community library of already-formalized mathematics (definitions, lemmas, and the standard
graph-theory vocabulary) that one builds on rather than re-deriving from scratch. Turning an informal,
natural-language statement into such a formal Lean statement is called
\emph{autoformalization}~\cite{wu2022autoformalization}; finding its proof is the job of an
\emph{automated theorem prover}. Recent \emph{neural} provers are large language models trained on
formal proofs whose goal is to output a candidate Lean proof---either in one shot (\emph{whole-proof}) or by
iterating against the compiler's error messages (an \emph{agentic} loop)---which is then handed to the
kernel for an independent check, so the model only proposes and the kernel disposes. There is a growing body of recent work on autoformalization and automated theorem proving especially in Lean, including for graph theory, see for instance \cite{wu2022autoformalization,mavani2026chipfiring,nader2026gtbench,kalfus2026ramsey,zhang2026leanmarathon,deepseek2025prover}. 

Although we recently found in \cite{davila2026graffiti3} certain experiments in this direction, to the best of our knowledge there is no existing system that integrates automated conjecturing, refuting, formalizing, and proving in a closed loop for discovery in graph theory.

In this paper, we describe and report on the development of such a computational system \textsf{AutoGraphForge} that aims to combine (i)-(iv) in a pipeline. The first two stages are combined into a \emph{generate-refute} loop, where conjectures are generated and then tested against a database of known theorems and (extremal) graphs with calculated invariants. Counterexamples to proposed candidate conjectures are being added to the dataset for future refinement of conjectures in rounds. The conjectures which survive this process are translated into a formal language Lean and then attempted to be proved using automated theorem provers.

Our current contributions are:

\begin{enumerate}
  \item We build the first version of the computational pipeline \textsf{AutoGraphForge} and run its generate--refute loop at scale on a high-performance computing
  cluster: a multi-round counterexample-guided loop partitioned into five full-node SLURM jobs, followed by a merge. The formalization-and-proving
  stage is implemented and integrated into the
  package (deterministic Lean export plus two state-of-the-art neural provers,
  \textsc{DeepSeek-Prover-V2-671B} served with vLLM and the Lean-specialised \textsc{OProver-32B}, with
  independent kernel verification against \texttt{mathlib4}); it passes initial sanity checks---the deterministic Lean export yields type-correct goals, and the
  prover independently kernel-verifies trivial
  inequalities ---but it has not yet been run as a controlled
  evaluation over the surviving conjectures (\S\ref{sec:formalize-prove}).
  \item We create a novelty filter of $559$ classical, folklore, and trivial relations
  (\S\ref{sec:heuristics_related})---spanning the chromatic, connectivity,
  matching/cover, domination, and zero-forcing families---closed under transitive composition of $f\le g$
  relations and under linear identity substitution. The filter also includes a linear-programming novelty test that decides whether a candidate inequality is implied by known relations.
  \item We build a refutation dataset of $348{,}207$ graphs (carrying up to
  $59$ exactly-computed invariants), combining the complete House of Graphs (HoG)~\cite{coolsaet2023house} export and many other extremal graph families, including the strongly-regular, minimal-Ramsey,
  Cayley, cage, cographs, minimally-rigid graphs, barbells, lollipops, spiders, etc., together with graphs generated from random models of regular, Erd\H{o}s--R\'enyi, bipartite, and tree graphs. 
  \item We implement counterexample-search algorithms 
    (\S\ref{sec:engine}) that include among others recent state-of-the-art methods such as probabilistic search using the cross-entropy method and Monte Carlo methods.
\end{enumerate}

We note that our system is not built from scratch but is built upon existing packages, frameworks and methodologies present in the state-of-the-art literature on each of the three areas of conjecturing, refuting, formalizing and proving. In particular, the conjecture generation is built on top of TxGraffiti2 \cite{davila2024automated} and Graffiti3 \cite{davila2026graffiti3}; the counterexample search algorithms are built upon recent state-of-the-art methods described in \cite{jooken2025computer,wagner2021constructions}; and the formalization and proving layer is built upon recent autoformalization methods and automated theorem provers, specifically Lean~4 with \texttt{mathlib4} as the target language and the \textsc{DeepSeek-Prover-V2}~\cite{deepseek2025prover} and \textsc{OProver-32B}~\cite{oprover2026} automated provers.

We are explicit that the loop is not yet closed: the first three stages run end-to-end at scale, and the proving stage is implemented and sanity-checked but not yet properly evaluated.

\section{Preliminaries}
\label{sec:preliminaries}

All graphs in this paper are considered finite and simple. For a graph $G$ we write $V(G)$ for its vertex set, $E(G)$ for its edge set, $n=|V(G)|$ for its \emph{order} (number of vertices) and $m=|E(G)|$ for its
\emph{size} (number of edges).  Our pipeline computes $59$ invariants in total for all the graphs, therefore to improve readability, we define only those that appear in our arguments and we
refer the reader to the \textsc{GraphCalc} documentation~\cite{davila2025graphcalc} for the exact definition of all 59 invariants. For the ones that do not appear in our main arguments but are mentioned, we use standard invariant symbols throughout the text: maximum and minimum degree $\Delta,\delta$, average degree $2m/n$;
independence number $\alpha$, matching number $\nu$,
vertex-cover number $\tau$, clique number $\omega$, chromatic number $\chi$,
domination number $\gamma$, and independent-domination
number $i$; vertex- and edge-connectivity $\kappa$ and $\lambda$; radius
$\mathrm{rad}$, diameter $\mathrm{diam}$, and triangle count $\mathrm{tri}$; and
the largest adjacency eigenvalue, or \emph{spectral radius}, written $\rho$
(not to be confused with the edge-connectivity $\lambda$). On a regular graph $\rho$ equals the common degree, giving the identity
$\Delta=\delta=2m/n=\rho$. The clique-cover number---the minimum number of
cliques covering $V(G)$, equivalently $\chi(\overline G)$---is written
$\overline{\theta}$ throughout. 

Beyond these classical invariants, our runs range over a family of less standard invariants that
will matter for the proving stage (\S\ref{sec:formalize-prove}), as they lie outside the provers'
training distribution. The \emph{zero-forcing number} $Z(G)$~\cite{aim2008zeroforcing} is the minimum size of a vertex set
$S\subseteq V(G)$ that colours all of $V(G)$ under the \emph{colour-change rule}: a coloured vertex
with exactly one uncoloured neighbour forces that neighbour to become coloured, applied until no
further forcing is possible. Its total analogue restricts the initial set $S$ to induce a subgraph with no isolated vertices, giving the \emph{total zero-forcing number}. 

We also use a handful of less common graph
classes in hypothetical conjectures---among them cographs ($P_4$-free graphs), split graphs (vertices can be partitioned into a clique and an independent set), and well-covered graphs
(every maximal independent set has the same size).

\section{The AutoGraphForge Architecture}

\subsection{System overview}
\label{sec:pipeline-fig}
\begin{figure}[htbp]
    \centering
    \includegraphics[width=\textwidth,keepaspectratio,trim=300 0 300 0,clip]{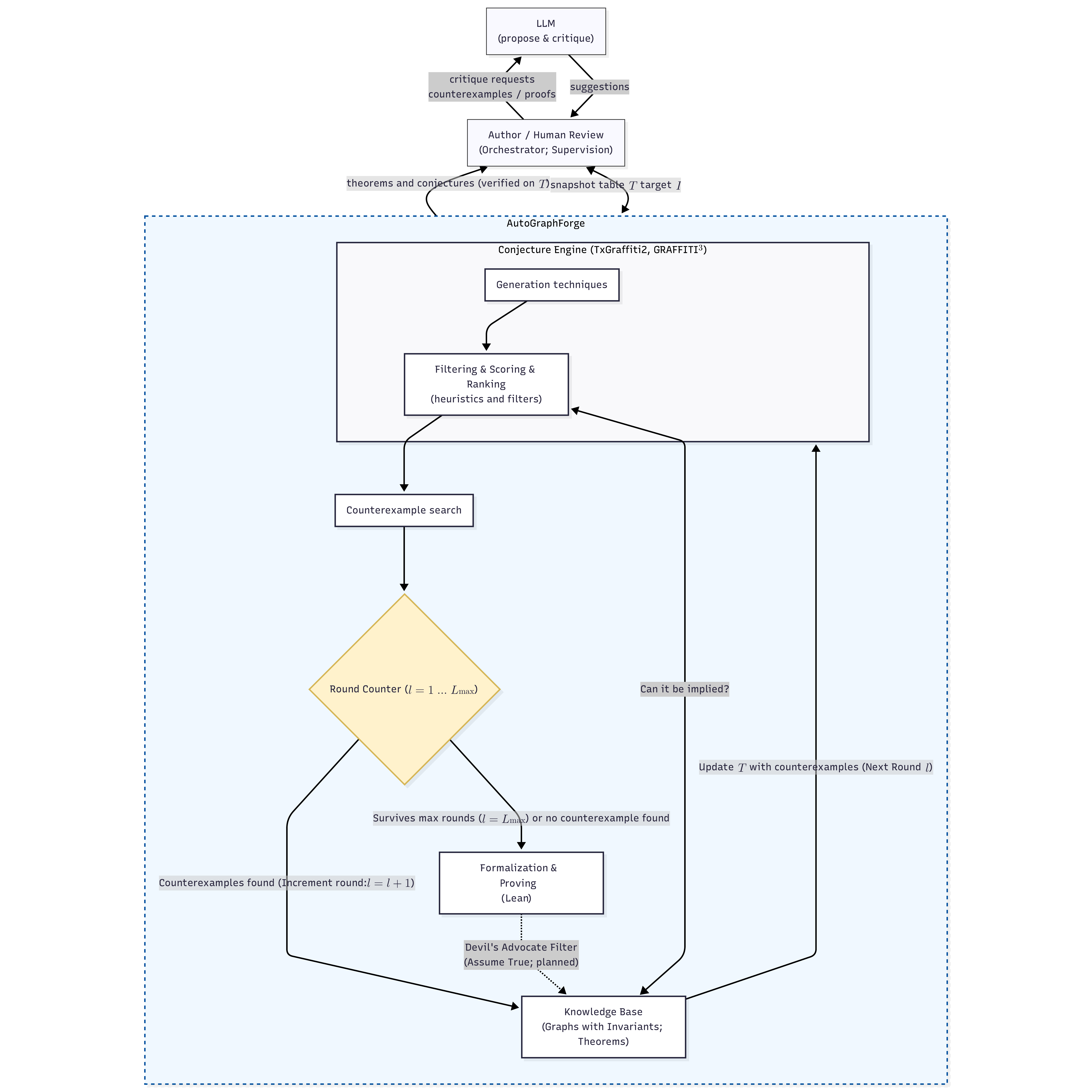}
    \caption{The complete AutoGraphForge pipeline. The orchestrator loop (LLM and Human) interacts with the AutoGraphForge and conjecture engine. Surviving conjectures enter a refutation loop managed by a round counter ($l = 1 \dots L_{\max}$), eventually proceeding to the formalization-and-proving stage if they survive $L_{\max}$ rounds. Conjectures are generated, immediately filtered and sorted with heuristics,
    and then aggressively tested by the counterexample searches (datasets with precomputed invariants and counterexample search algorithms). Only conjectures surviving
    multiple rounds of refutation are passed to the deterministic Lean export and the
    kernel-checked neural provers (\S\ref{sec:formalize-prove}; found
    counterexamples are inserted into the knowledge
    base. The dashed line is the planned Devil's Advocate filter (future work; see
    the Conclusion), where surviving
    conjectures would be assumed true to constrain future generation of candidate conjectures.}
    \label{fig:pipeline}
\end{figure}

At a high level, \textsf{AutoGraphForge} is a system
(Figure~\ref{fig:pipeline}) that couples generation of conjectures, selection heuristics,
refutation search, and a formalization/proving layer, in the spirit of the proposed
competing optimist/pessimist ``GraphMind'' architecture ~\cite{davila2024optimist}. The \emph{knowledge base} consists of the current snapshot table $T$ of graphs together with their computed invariants and the library of known theorems used by the novelty filter (\S\ref{sec:heuristics_related}). We treat the example and counterexample graphs as the rows of this table $T$,
whose numerical columns embed the graphs in $\mathbb{R}^k$ (where $k$ is the number of invariants).

Candidate conjectures are generated based on $T$, filtered, scored and ranked, then aggressively tested by the refutation engine while  every counterexample is inserted to $T$ for the next round of generation of candidates; only statements that survive multiple rounds advance to
autoformalization, proving, and human review. The remaining sections detail each stage in turn: generation, filtering and sorting of conjectures (\S\ref{sec:theory-selection}),
refutation (\S\ref{sec:refute}), and formalization and proving
(\S\ref{sec:formalize-prove}).

\paragraph{Validation of invariant calculation.} We use
\textsc{GraphCalc}~\cite{davila2025graphcalc} to calculate invariants and
every value that is stored is its exact computation, or left empty if timed out. An
independent \texttt{networkx}~\cite{hagberg2008networkx} list is registered purely as a cross-check, and
a \texttt{crossvalidate} script compares the two on sample graphs --- this is implemented as unit test before every run. Further, we verified computation of invariants of \textsc{GraphCalc} against the independently-computed HoG values on $2{,}500$
graphs: for every shared invariant
($\chi,\alpha,\omega,\gamma,\nu,\Delta,\delta,\kappa,\lambda,\mathrm{diam},
\mathrm{tri}$, and the booleans) there were zero mismatches.

\subsection{Conjecture generation as theory selection}
\label{sec:theory-selection}

\textsf{AutoGraphForge} takes \textsc{TxGraffiti2}~\cite{davila2024automated} and \textsc{Graffiti3}~\cite{davila2026graffiti3} as its foundation. The canonical representation of a
conjecture is TxGraffiti2's own \texttt{Conjecture} object, and every subsequent component described in the next subsections---the implication-deciding novelty filter, the
selection heuristics, and the refutation engine---is built to take these
native objects directly. 

Here we give a concise overview of the conjecture generation and selection process, which is described in detail in \cite{davila2024automated,davila2026graffiti3}.  A linear grammar of conjectures consists of halfspaces
$\sum_i a_i f_i(G)\le c$ (with $a_i, c \in \mathbb{R}$) that must hold for every row of the current snapshot $T$.
The convex hull of the cloud makes the core difficulty geometric: there are
infinitely many inequalities true on the current snapshot $T$ (any halfspace whose boundary lies
above the hull), but only the \emph{facets}---tight on at least one extremal
object---carry information, and no single inequality is ``the best.''  

We build upon TxGraffiti2's native generators and its implementation of the Dalmatian dominance filter further specified in \cite{davila2024automated,davila2026graffiti3}. Invariants
are supplied by \textsc{GraphCalc} (\S\ref{sec:data}). In the runs reported here the generator is
\textsc{Graffiti3}~\cite{davila2026graffiti3} (the conjecturing
engine of the TxGraffiti package) which additionally can generate ratio,
product, square roots and $\log$ bounds. Beyond numeric inequalities, the engine operates in a boolean mode using the \emph{Sophie} heuristic. Formally, let $\mathcal{P} = \{P_1, P_2, \dots, P_m\}$ be a dictionary of computable boolean predicates on graphs (e.g., regularity, planarity, claw-freeness, chordality, $3$-connectivity, vertex transitivity). Given a target property $I$, Sophie shifts from producing a numeric bound to generating \emph{sufficient conditions} by searching the space of conjunctions $H = \bigwedge_{i \in Q} P_i$ (where $Q \subset \{1, \dots, m\}$ and $k = |Q| \ge 1$) for statements of the form $H \implies I$. A candidate is retained if $H(G) \implies I(G)$ holds for all graphs $G$ in the dataset. Moreover, the system can also search for \emph{necessary conditions}. This is all generated on current snapshot $T$ and its computed $59$-invariants. 

Generation of conjectures begins from \emph{expressive graphs} (a curated default set that together with all $59$-invariants computed that forms the initial snapshot $T$) available in TxGraffiti package and grows only by counterexamples.

\subsection{Sorting candidates and novelty filtering}
\label{sec:heuristics_related}
Without aggressive filtering, conjecture systems suffer a ``combinatorial
explosion,'' generating thousands of trivial or redundant statements. 
% \emph{Simplicity Over Complexity principle}: While optimization models can handle bounds with many invariants simultaneously, based on an Ockham's razor principle, bounds with both fewer invariants, and arithmetic and logical operations are preferred. 
Therefore, it is required that the system assert only the
tightest bound for which no counterexample is known; this is the formal basis of
the Dalmatian dominance filter and the polyhedral facet approach alike.
\textsc{Graffiti3}~\cite{davila2026graffiti3} filters and sorts candidate bounds by applying three primary sorting heuristics originating from the \textsc{Graffiti} lineage: the \emph{Dalmatian}, \emph{Morgan}, and \emph{Touch} heuristics. The \emph{Dalmatian heuristic} (dominance), due to Fajtlowicz and revisited by Larson and Van~Cleemput~\cite{fajtlowicz1988conjectures,larson2016dalmatian}, discards a candidate whenever some accepted conjecture is never weaker and sometimes strictly tighter. The \emph{Morgan heuristic} drops a bound that is asserted on a smaller, restricted graph class when the exact same bound also holds on a broader superclass containing it. Finally, the \emph{Touch heuristic} ranks candidates by the number of times they are sharp (or tight) on the dataset graphs. Conjectures with a higher touch count more closely trace the extremal boundary of the dataset and are highly prioritized. These three classical heuristics act as the primary filters used throughout our pipeline.

Discovering a candidate inequality is only part of the task: the system must also
decide whether the inequality is genuinely new or merely a combination of
already-known relations. We propose the following for linear candidate inequalities.

Suppose the system generates a candidate inequality
    \[
        f \le R(g),
    \]
    where $f$ is a target invariant (for example, the independence number) and $R(g) := \sum_i^n c_i g_i + c_0$ is a linear combination of other invariants $g_i$, with $c_i, c_0 \in \mathbb{R}$ and $g := (g_1, g_2, \ldots, g_n)$. Before accepting it, the system tests it against a knowledge base which contains known theorems that bound the
  same target invariant ($f \le B_1(g)$, $f \le B_2(g)$, and so on, where each $B_j(g)$ is a known linear bound on $f$). Crucially, to preserve soundness, the system only admits a bound $B_j(g)$ into the aggregate if it is valid over
  the candidate conjecture's entire hypothesis class (or a superclass of it).
    
    The candidate is declared known (redundant) if it is implied by a convex combination of these theorems. Assigning a non-negative weight $w_j$ to each $B_j(g)$ with $\sum_j w_j = 1$ yields the aggregate bound $f \le \sum_j w_j B_j(g)$. If this aggregate is at least as tight as $R(g)$, the
  candidate is redundant; equivalently, the difference between the candidate and the aggregate is non-negative everywhere:
    \[
        R(g) - \sum_j w_j B_j(g) \ge 0.
    \]
    
    \subsubsection*{The non-negativity certificate}
    
    To certify that this difference is non-negative, we express it identically as a sum of terms that are individually non-negative. We seek weights $w_j \ge 0$, non-negative slack variables $s_i, t \ge 0$, and free multipliers $\mu_k \in \mathbb{R}$ such that the following identity holds for all $g$:
    \[
        R(g) - \sum_j w_j B_j(g) = \sum_i s_i(g_i - L_i) + \sum_k \mu_k E_k(g) + t.
    \]

    The right-hand side is composed of terms each guaranteed to be non-negative:
    \begin{itemize}
        \item \emph{Safe lower bounds} $s_i(g_i - L_i)$: every invariant $g_i$ has a known lower bound $L_i$, so $(g_i - L_i) \ge 0$; multiplying by a non-negative
multiplier $s_i$ preserves non-negativity. As with the tabulated theorems,
each bound $g_i \ge L_i$ is admitted only if it holds unconditionally over the
candidate's entire hypothesis class (or a superclass of it); global bounds
valid on all graphs are admissible for every candidate.
        \item \emph{Class identities} $\mu_k E_k(g)$: linear structural relations $E_k(g)=0$ that hold identically on a graph class $P_k$, so multiplying by any free multiplier $\mu_k$ does not affect non-negativity. Crucially, soundness requires that $E_k$ strictly vanish on the whole domain the
  candidate is quantified over, so each identity is similarly restricted by graph class: it is admitted only when the candidate's hypothesis class is $P_k$ or a subclass of it. For candidate conjectures with no hypothesis class restriction, only universal identities (those with $P_k$ = all graphs, e.g.\ Gallai's
  $n=\alpha+\tau$) are available.
        \item \emph{Constant slack} $t$: a non-negative scalar absorbing the residual constant.
    \end{itemize}

    Because the invariants $g_i$ are treated as independent symbolic variables, matching the coefficient of each $g_i$ on both sides reduces the problem to a system of linear equations. Verifying redundancy therefore reduces to a linear-programming (LP) feasibility test: if the solver finds weights and
  slacks satisfying the identity, the candidate is provably a structural consequence of the tabled theorems and is \emph{flagged as a known rediscovery}---excluded from the set of \emph{novel} survivors, but stored, so it can serve as a soundness check (the $33$ flagged rediscoveries of \S\ref{sec:dynamics}).

We inserted into the novelty filter a list of known relations, including classical theorems, folklore, and trivial bounds from standard books and literature in graph theory. Together the list holds $559$ entries after closing the simple $f\le g$
relations under transitivity. For instance the following are captured in the list: Whitney's connectivity chain $\kappa\le\lambda\le\delta$~\cite{whitney1932congruent}
(all graphs), K\"onig's $\nu=\tau$~\cite{konig1931graphen} (bipartite), and
Gallai's $n=\alpha+\tau$~\cite{gallai1959uber} (all graphs)---each applied only on the class where
it holds. The LP part correctly hides, for
example, $\kappa\le\tfrac12\delta+\tfrac12\lambda$.

The same LP-implication test decides conjectures in both directions of an
implication, not only single upper bounds. An \emph{equality} $f=g$ (under a
hypothesis $H$) is flagged known when \emph{both} $f\le g$ and $f\ge g$ are
implied by the table---i.e. the bound is simultaneously necessary and
sufficient (e.g. the Allan--Laskar theorem: claw-free $\Rightarrow \gamma=i$~\cite{allan1978domination}). Sophie \emph{necessary conditions} are handled by contraposition. Such a
condition has the form $A\Rightarrow\neg C$, where $A$ is a numeric predicate
and $C$ a graph class. It is flagged known whenever the contrapositive
$C\Rightarrow\neg A$ appears in the table. For instance
$(\chi>2)\Rightarrow\neg\textsf{bipartite}$ is recognised as the
contrapositive of the known bound $\textsf{bipartite}\Rightarrow\chi\le2$.
Class inclusions ($\textsf{tree}\subset\textsf{bipartite}$,
$\textsf{cubic}\subset\textsf{regular}$, \dots) are applied throughout, so a
subclass inherits every bound proved for its superclasses. In the \emph{sufficient} direction, a Sophie condition
$A\Rightarrow C$ with a positive class conclusion is flagged known only against
a hand-curated table of textbook \emph{characterizations} (e.g. $2m/n=\Delta
\Rightarrow \textsf{regular}$, or $a(G)>0\Rightarrow\textsf{connected}$ by
Fiedler); crucially these cannot be data-mined, since a rare invariant equality
can agree with a class on every test graph by coincidence without being a
theorem, so only genuine characterizations are admitted and all other
$A\Rightarrow C$ candidates are conservatively left novel.

\subsection{Counterexample search and refutation}
\label{sec:refute}
Every candidate conjecture that survives novelty and selection advances to this stage.
Because there are infinitely many non-isomorphic finite graphs and the current snapshot $T$ is only a partial view,
``true on the table'' is a statement about present evidence, not a
generalisation claim. Therefore, the system that tries to find robust conjectures must actively search for counterexamples. There are several approaches to automated refutation. For a review see \cite{jooken2025computer}. Recently, deep reinforcement learning~\cite{wagner2021constructions} has been applied for counterexample search, building counterexample graphs edge by edge guided by the conjecture's violation margin.

In our pipeline, we first check the candidate against the static dataset of $348{,}207$ graphs (\S\ref{sec:data}). Then, if no counterexample is found there, the conjecture is tested on some parametric families and graphs generated from random graph models. If the candidate survives, counterexamples are sought by 
specialised search algorithms (\S\ref{sec:engine}). When at any point counterexample is found, the counterexample is inserted into the snapshot $T$, changing the feasible
space of all subsequent conjectures in the next round. If no counterexample is found, the conjecture gets to the formalization and proving layer (\S\ref{sec:formalize-prove}).

\subsubsection{Datasets of potential counterexamples}
\label{sec:data}

The conjecture engine generates candidates over a small snapshot $T$ (a few hundred graphs initially; \S\ref{sec:dynamics}), but the refutation engine tests them against a much larger dataset. We combine
several sources. The HoG export~\cite{goedgebeur2022introduction,coolsaet2023house} provides $28{,}859$ graphs with $\sim\!28$
pre-computed invariants each (treewidth, feedback-vertex-set number, spectral
data, vertex cover, girth, \dots), parsed directly so that nothing is
recomputed; values marked \texttt{undefined}, \texttt{infinity}, or
\texttt{Computation time out} are treated as missing. The dataset also contains every connected graph on $2\le n\le 9$ vertices ($273{,}192$ graphs)
through an exact invariant list and a set of computed graph-class flags
(triangle-free, cubic, claw-free, cograph, split, outerplanar,
self-complementary, Hamiltonian, well-covered). We also used the HoG meta-directory to collect cograph, minimal-Cayley, cage,
minimal-Ramsey, strongly-regular (Spence), and minimally-rigid graph families. The
union contains $348{,}207$ graphs ($28{,}859$ from HoG, $273{,}192$ from the connected census on
$2\le n\le 9$, and $46{,}156$ from the special families obtained from HoG meta-directory). 

The dataset contains $59$ computed invariant
columns. These $59$ split into $45$ \emph{numeric} invariants ($\alpha,\nu,\chi,\gamma,\dots$) and $14$
\emph{boolean} graph-class predicates (triangle-free, claw-free, cubic, cograph, split, Hamiltonian,
\dots). Only the $45$ numeric invariants serve as the left-hand-side \emph{targets} of generated
inequalities (hence the ``$45$ numeric targets'' of \S\ref{sec:dynamics}), while the $14$ booleans---together with two
derived order thresholds ($n\ge 3$ and $n\ge 4$) that the battery adds precisely so that
bounds failing only on the smallest graphs become true class-restricted statements---give the
$16$ predicates available to Sophie as hypotheses and class conditions, not as numeric targets. This static
dataset is one refutation resource. Then, there are two further datasets on which the conjecture is tested computed from parametric families (barbells, lollipops,
spiders, $k$-trees, line graphs, Delaunay graphs, $\alpha=2$ complements, \dots). A finite dataset usually does not certify a universally-quantified statement. Thus if it makes sense to do so, the pipeline tests candidates on random graph models (random-regular, Erd\H{o}s--R\'enyi $G(n,p)$, random trees,
and random bipartite graphs, several per (class,\,order)), so that refutation probes the average behavior corresponding to these graph classes.  We note that these datasets of graphs are used exclusively for refutation and note that conjecture generation never runs over the full dataset. Instead generation of conjectures always operates on a small initial snapshot $T$---the $337$ TxGraffiti expressive graphs, a few hundred---that grows only by counterexamples and persists across runs: by the time the experiments of \S\ref{sec:dynamics} begin, $T$ has already grown, over earlier development runs predating this report, to $2{,}860$ graphs (\S\ref{sec:dynamics} makes this pre-run growth explicit before reporting the experiments proper), described in \S\ref{sec:theory-selection}. Generating over the whole dataset would be computationally prohibitive and, more importantly, a small counterexample-hardened core suffices to drive generation.

\subsubsection{Counterexample-search algorithms}
\label{sec:engine}
\label{sec:backends}

For conjectures whose counterexamples lie outside our fixed datasets, constructions and graphs generated by random models, the system
provides counterexample-search algorithms, all sharing a single
margin interface. The \emph{margin} of a graph $G$ against
a candidate measures how badly $G$ violates it: for an inequality conjecture $f(G)\le R(g)(G)$ it is the
amount of violation
\[
  \mathrm{margin}(G) \;=\; f(G) - R(g)(G),
\]
and for a Sophie sufficient-condition $H\Rightarrow I$ it is positive exactly on a graph that
satisfies the hypothesis $H$ but fails the consequent $I$. Thus $\mathrm{margin}(G)>0\iff G$ refutes the
candidate, and a larger margin means a more decisive counterexample, so every backend below is simply a
different way of maximising the margin over graphs. The
backends are:

\begin{itemize}
  \item \emph{SMT (satisfiability-modulo-theories) encoding} (\textsf{z3}): This method translates the search for a counterexample into a logical formula for specialized solver. Graph structure is encoded as Boolean edge variables; invariant constraints ($k$-colorability, independence set size, etc.)\ are expressed as pseudo-Boolean formulae and dispatched to a Z3~\cite{demoura2008z3} solver.
  \item \emph{Linear cross-entropy} (\textsf{cross\_entropy}): Intuitively, this algorithm maintains a probabilistic generative model of a graph, samples candidate graphs, and updates the model parameters to maximize the likelihood of the highest-performing candidates. It is a gradient-free distribution search~\cite{wagner2021constructions}: a per-edge inclusion probability matrix is sampled in batches, the elite fraction of graphs with the largest margin is retained, and the edge probabilities are nudged towards the elite frequencies.
  \item \emph{Variable-neighbourhood search} (\textsf{vns}): This procedure attempts to improve the objective via local edge modifications, applying increasingly larger stochastic perturbations to escape local optima when the ascent stagnates. It alternates between a $k$-edge flip ``shake'' move and a greedy $1$-edge local ascent; if the resulting graph improves the conjecture's margin, the neighbourhood size resets to $k=1$, otherwise $k$ increments.
  \item \emph{Monte-Carlo tree search} (\textsf{mcts}): The technique builds a decision tree of possible edge modifications, carefully balancing the exploration of untested graph structures against the exploitation of known promising pathways. We use an Upper-Confidence-Bound bandit search over edge addition and deletion actions, with leaf values estimated by random edge-flip rollouts that return the maximum margin observed.
  \item \emph{Simulated annealing} (\textsf{sa}): Inspired by metallurgical annealing, the search frequently accepts non-improving edge modifications in the early stages to broadly explore the search space, but becomes increasingly strict over time to hone in on a solution. It performs classical simulated annealing over local edge modifications under a linear cooling.
  \item \emph{Deep reinforcement learning} (\textsf{rlgt-RL}): This method models the search as a Markov decision process where an artificial neural network policy is trained to maximize a reward signal based on the conjecture's violation margin. Implemented via the RLGT framework~\cite{damnjanovic2026rlgt} (building on~\cite{wagner2021constructions}), it trains a neural edge-selection policy (using Deep Cross-Entropy) guided by the reward which is higher if the graph is closer to refuting the conjecture and lower otherwise.
\end{itemize}

More comprehensive reviews of these methods are in~\cite{jooken2025computer,wagner2021constructions}. In our pipeline, the backends above can be initialised randomly from scratch or seeded from another graph.

\subsection{Formalization and proving}
\label{sec:formalize-prove}
Surviving conjectures that pass refutation are handed to the formalization and proving layer. This stage
is implemented and wired into the package, but---unlike the generate--refute loop---it has not yet been fully
run over the surviving conjectures; so far it has only undergone initial sanity checks on trivial
conjectures, which it passed (below). We describe our progress and the planned evaluation.

Because a TxGraffiti conjecture is already a formalized object---a relation between named invariants
under a conjunction of graph-class predicates---translating it to Lean~4~\cite{moura2015lean} is a deterministic translation
(\texttt{pipeline/lean\_export.py}). Each invariant column maps to its Lean name (e.g.\
\texttt{zero\_forcing\_number}\,$\mapsto$\,\texttt{G.zeroForcingNumber}), each class predicate to a
preamble hypothesis (\texttt{nontrivial}\,$\mapsto$\,\texttt{G.IsNontrivialClass}), and the relation
symbol to itself; binders are injected and both sides are cast to $\mathbb{R}$. The resulting statement is
faithful by construction---the only trust assumption is that the preamble definitions match the
intended invariants---and any conjecture using an invariant or class outside the formalized tables is
skipped. For example, 
$(\textsf{nontrivial})\Rightarrow\textsf{zero\_forcing\_number}\le\textsf{total\_zero\_forcing\_number}$
is translated mechanically as
\begin{quote}\small\ttfamily\noindent
theorem CEGIS\_1 \{V : Type*\} [Fintype V] [DecidableEq V]\\
\null\quad(G : SimpleGraph V) [DecidableRel G.Adj] (\_h0 : G.IsNontrivialClass)\\
\null\quad: (G.zeroForcingNumber : $\mathbb{R}$) $\le$ (G.totalZeroForcingNumber : $\mathbb{R}$) := sorry
\end{quote}
The output is a \texttt{sorry}-terminated skeleton that the Lean kernel elaborates to a type-correct goal,
not a certified proof. For proving we integrate two
state-of-the-art open-weight neural provers: \textsc{DeepSeek-Prover-V2-671B}, served with
vLLM~\cite{kwon2023vllm} in tensor-parallel across eight H200 GPUs, and the Lean-specialised
\textsc{OProver-32B}~\cite{oprover2026} (a Qwen3-based whole-proof prover, state-of-the-art on MiniF2F
among open-weight provers), each usable in single-shot whole-proof mode and in a multi-round
agentic mode which has feedback from compiler. Soundness is enforced
independently of the models: every candidate proof is kernel-checked against the
\texttt{mathlib4}~\cite{vandoorn2020maintaining} library and our \texttt{GraphInvariants} preamble and each proof that passes is persisted as a self-contained \texttt{.lean} file and re-verified by a standalone kernel pass, so any count we eventually
report will be reproducible from these files.

Two preliminary checks confirm that the stage's components function end-to-end, short of a full
evaluation. First, the deterministic export is faithful and type-correct: the supported survivors
translate to Lean~4 goals that the kernel accepts modulo a \texttt{sorry} placeholder, so the
obstacle to closing the loop is proof search, not statement elaboration. Faithfulness here is by
construction (above), but it rests on the preamble definitions being correct; the author therefore
manually checked the formalized invariant definitions and a sample of the rendered goals, confirming they
encode the intended conjecture. 

The proof stage is also ready. \textsc{OProver-32B} model produced proofs of two trivial inequalities over standard
\texttt{mathlib} invariants---$\delta\le\Delta$ (minimum degree at most maximum degree) and $\omega\le n$
(clique number at most order)---that pass the independent kernel check and were persisted and re-verified
as self-contained \texttt{.lean} files. These are deliberately trivial and they establish that the deterministic export, the prover interface, and the verification can be integrated together. 

We expect proving to be the hardest stage. One likely reason is that many surviving conjectures rely on graph invariants such as zero-forcing are defined
only in our preamble and absent from the datasets these provers were trained on.

\section{Discovery in graph theory}
\label{sec:discovery}

We now report on the iterative process of the pipeline and its preliminary outcomes.

\subsection{Iterative run dynamics on the cluster}
\label{sec:dynamics}

The snapshot $T$ persists across all development and reported runs, so it is not the bootstrap
$337$ TxGraffiti expressive graphs by the time we report numbers below---it already carries witnesses
accumulated over earlier runs. To establish a baseline, we first ran a single-pass generation
(one generate--refute round) from an intermediate snapshot of $1{,}554$ graphs ($337$ expressive plus
$1{,}225$ witnesses accumulated by that point, unioned and deduplicated); it yielded $3{,}959$ candidates, of which $1{,}249$ were
refuted. The datasets dominated the rejections ($865$ by the extremal families, $217$ by graphs generated from the random
models, and $161$ by the House of Graphs dataset). The rest were refuted by counterexample
search algorithms. This validates the strategy of testing against fast, parametrised datasets before
resorting to expensive counterexample search.

Building on this baseline, we ran \textsf{AutoGraphForge} as five independent partitions, each on a
full $256$-core CPU node, that together partition the $45$ numeric target invariants; the
partitioning scheme, cluster hardware, and the $1.22$ CPU-year budget are detailed
in Appendix~\ref{app:compute}. By this point further development runs had grown the persisted witness
set further, so each partition started from a common snapshot $T$ of
$2{,}860$ graphs (the same $337$ TxGraffiti expressive graphs plus $2{,}531$ accumulated witnesses,
unioned and deduplicated) and ran the generate--refute loop until no new counterexamples were found.
In short, $T$ grew $337\to1{,}554\to2{,}860$ across these successive runs, before the round-by-round
growth reported in Table~\ref{tab:rounds} below, which summarises all five partitions, labelling each
by the span of its nine invariant targets.

\begin{table}[!h]
  \caption{Per-partition iterative dynamics (all five partitions). The snapshot $T$ is
  the small evolving set of graphs that conjecture generation runs over (not the
  full $348{,}207$-graph refutation dataset); all partitions start from a common snapshot $T$ of
  $2{,}860$ graphs and grow it only by witnesses. Columns R1--R4 and ``$\ge\!5$''
  give the new counterexamples added back into the snapshot $T$ in each round (only partition
  1 ran past round~4); ``Total'' is their sum, so the final size of $T$ is $2{,}860+\text{Total}$.
  A round adding zero witnesses is a fixed point. ``Surv.'' is the partition's final survivor count;
  the five sum to the $8{,}281$ raw survivors. The merge step (\S\ref{sec:dynamics}) reduces these to
  $6{,}522$ in two stages: $1{,}413$ are refuted by witnesses found in another partition---each
  partition grows its own hard seed and never sees the others', so a graph that kills a candidate
  may sit in the wrong shard---and a further $346$ are cross-partition duplicates.}
  \label{tab:rounds}
  \resizebox{\columnwidth}{!}{%
  \begin{tabular}{lrrrrrrrr}
    \toprule
    Partition (each consisting of $9$ targets) & Rds & R1 & R2 & R3 & R4 & $\ge\!5$ & Total & Surv. \\
    \midrule
    0 (algebraic connectivity\,--\,diameter) & 5 & $793$ & $164$ & $21$ & $12$ & $0$  & $990$ & $1{,}762$\\
    1 (domination number\,--\,maximum degree) & 9 & $499$ & $56$  & $11$ & $24$ & $25$ & $615$ & $1{,}438$\\
    2 (min.\ maximal matching\,--\,residue) & 5 & $494$ & $39$  & $9$  & $1$  & $0$  & $543$ & $1{,}815$\\
    3 (Roman domination\,--\,total domination) & 5 & $562$ & $66$  & $11$ & $6$  & $0$  & $645$ & $1{,}345$\\
    4 (total zero forcing\,--\,zero forcing) & 5 & $580$ & $48$  & $3$  & $1$  & $0$  & $632$ & $1{,}921$\\
    \bottomrule
  \end{tabular}}
\end{table}

Three patterns are robust across partitions. (i) The first round alone contributes
$\sim\!500$--$790$ new witnesses (a $\sim\!20$--$28\%$ expansion of $T$), after which the
per-round yield falls sharply and trends toward zero---$\sim\!40$--$160$ in
round~2 and $\sim\!3$--$20$ in round~3 (Table~\ref{tab:rounds}). The decline is
not strictly monotone, however: partition~1 rebounds in its later rounds
($11$ witnesses in round~3, then $24$ in round~4 and $25$ in later rounds) as a delayed
cluster of counterexamples surfaces for its domination/degree targets, before it
too reaches a fixed point---a round that adds zero witnesses, at which the
snapshot $T$ stops growing and the loop terminates---at round~9; the other four
partitions terminate at round~5. \emph{(ii) In this run, counterexample witnesses are dominated by the
dataset; the counterexample search algorithms are surprisingly
a minor contributor.} The entire set of algorithms (z3 / cross-entropy / VNS / SA
/ MCTS / RL) accounts for only $1$--$22$ counterexamples per round. We note this is a
single-run observation over stochastic searchers run at default hyperparameters (configurations
are in the public repository), so it should not be read as a general verdict on the methods; a possible
reason is precisely that we did not optimize their hyperparameters, and we plan to investigate this further.

Merging the five partitions' output conjectures ($8{,}281$ in total) and removing duplicate
statements yields $6{,}522$ survivors. The class-conditioned inequalities
are disjoint across partitions (each partition owns distinct target invariants on the
left-hand side), but the Sophie sufficient-conditions are not: their
consequent or antecedent is a target-independent boolean property, so different partitions might
re-derive the same condition, and removing duplicates eliminates $346$ such cross-partition
repeats. The survivors split into $2{,}677$ class-conditioned inequalities and
$3{,}845$ Sophie conditions. Essentially none of the surviving novel candidates is an
unconditioned inequality: the unconditioned bounds our generator proposed were either eliminated
by disconnected graphs in the refutation dataset (which break bounds that tacitly assume connectivity) or
recognised as known, leaving the survivors in their class-conditioned forms. This is not a claim that
unconditioned inequalities cannot hold in general---$\Delta\ge\delta$, for instance, holds for every
graph---only that, in this run, the surviving novel candidates were class-conditioned. The novelty filter
flags $33$ survivors as rediscoveries of classical results---among them Gallai's identity $\tau = n - \alpha$, the bound $\mathrm{rad} \le \alpha$ (conjectured by \textsc{Graffiti} and proved by Erd\H{o}s, Saks, and S\'os~\cite{erdos1986induced}), the K\"onig--Egerv\'ary theorem $\nu = \tau$, the perfect-graph identity $\alpha = \overline{\theta}$, and fundamental inequalities such as $\nu \le \tau \le 2\nu$, $\nu \le n/2$, and $\gamma \le i(G)$---each recovered independently on the chordal, cograph, and bipartite classes. We read these rediscoveries as a soundness check: the pipeline recovers known theorems and correctly does not flag them as novel. We note that they are expected---the generator recombines invariants, so producing known relations is the design working, not a discovery---and therefore evidence correctness, not discovery capability; the genuine test is whether the novel survivors are meaningful, which we begin to examine in \S\ref{sec:worked}. 

\subsection{Exemplary machine-generated discoveries}
\label{sec:worked}

To probe whether the surviving conjectures are tractable and meaningful rather than syntactic noise, we
hand-examined the simplest ``novel''-flagged inequalities with the highest touch numbers. We manually checked the top $100$ survivors by touch number. Many ``novel''-flagged survivors that we checked are, despite the heuristics and novelty filter, dominated by
\emph{true-but-trivial} bounds and by \emph{definitional artefacts}---such as the Sophie condition
$(\mathrm{radius}\le k)\Rightarrow\textsf{connected}$ for $k\ge 1$, whose hypothesis is
only defined on connected graphs---that our novelty filter does not yet recognise.

To replace that impression with a count, we classified all $100$ inspected survivors
automatically (\texttt{tools/audit\_sample.py}); the categories are computed from the
pipeline's own novelty table, subsumption lattice and support statistics rather than
assigned by hand, so the classification is reproducible from the released artefacts
(Table~\ref{tab:audit}). Just over half are triage successes in the pessimistic sense:
$49$ are rediscoveries the novelty filter recognises, $2$ are logically implied by a
stronger survivor already in the list, and $1$ is \emph{decorative}---its class
hypothesis is satisfied frame-wide and so does no work. The remaining $48$ are not
known to us to be either trivial or false. We deliberately do not report a
``probably false'' category: every item here survived refutation against the full
$348{,}207$-graph refutation dataset, so the pipeline has no evidence of falsity for any of them,
and asserting one would be an unsupported guess.

\begin{table}[!h]
  \caption{Automatic classification of the $100$ highest-touch survivors.}
  \label{tab:audit}
  \begin{tabular}{lr}
    \toprule
    Category & Count \\
    \midrule
    Rediscovery recognised by the novelty filter & $49$ \\
    Subsumed by a stronger survivor & $2$ \\
    Decorative hypothesis (class does no work) & $1$ \\
    Promising, universal (no class hypothesis) & $28$ \\
    Promising, class-conditioned & $20$ \\
    \midrule
    Total & $100$ \\
    \bottomrule
  \end{tabular}
\end{table}

However, we also found some conjectures that are nontrivial, true, and provable by a short argument relating the \emph{annihilation number}
$a(G)$ (the largest $k$ such that the $k$ smallest degrees sum to at most $m$) and the \emph{edge-cover
number}, the minimum number of edges covering $V(G)$, which we write $\mathrm{ec}(G)$ to avoid clashing
with the spectral radius $\rho$. Two class-conditioned survivors form a complementary pair---the bound
flips direction between the bipartite class (rank $91$ by touch, touch $922$) and the regular
class (rank $177$, touch $589$). Ranks are among the $2{,}677$ survivors that carry a
recomputed, hypothesis-aware touch count; the cubic specialisation (rank $453$, touch $233$)
is now flagged as subsumed by the regular form, since \textsf{cubic}$\subset$\textsf{regular}:
\begin{align}
  G \text{ is bipartite} &\;\Rightarrow\; \mathrm{ec}(G) \le a(G), \label{eq:bip}\\
  G \text{ is cubic (more generally regular)} &\;\Rightarrow\; a(G) \le \mathrm{ec}(G). \label{eq:cub}
\end{align}
Neither is in our novelty table, yet both are provable by combining classical results. Neither of these is a deep new theorem. Nevertheless, they show the pipeline works---it can produce true,
refutation-hardened statements that reduce to a short, human-checkable argument rather than to noise. 

\smallskip\noindent\textit{Proof of \eqref{eq:bip}.}
Assume $G$ has no isolated vertices. Gallai's identities give $\alpha+\tau=n$ and $\nu+\mathrm{ec}=n$, so
$\mathrm{ec}=n-\nu$; K\H{o}nig's theorem on bipartite graphs gives $\nu=\tau$, whence
$\mathrm{ec}=n-\tau=\alpha$. Pepper's bound $\alpha(G)\le a(G)$ (introduced in his 2004 dissertation~\cite{pepper2004thesis}; see also~\cite{pepper2009}) then yields
$\mathrm{ec}(G)=\alpha(G)\le a(G)$. (For a general graph $\nu\le\tau$ gives only $\mathrm{ec}\ge\alpha$, so
bipartiteness is essential to the equality $\mathrm{ec}=\alpha$ and hence to the bound.) \hfill$\square$

\smallskip\noindent\textit{Proof of \eqref{eq:cub}.}
Let $G$ be $r$-regular on $n$ vertices, so $m=rn/2$. The $k$ smallest degrees sum to $rk$, and $rk\le
rn/2 \iff k\le n/2$, so $a(G)=\lfloor n/2\rfloor$. Since a matching covers two vertices per edge,
$\nu(G)\le n/2$, and Gallai gives $\mathrm{ec}(G)=n-\nu(G)\ge n/2\ge a(G)$. Equality $a(G)=\mathrm{ec}(G)$
holds exactly when $\nu(G)=n/2$, i.e.\ when $G$ has a perfect matching---so for bridgeless cubic graphs
the bound is tight (by Petersen's theorem), while a cubic graph without a perfect matching satisfies it
strictly. \hfill$\square$

\section{Discussion and future work}
\label{sec:conclusion}

Further analysis of the $6{,}522$ surviving conjectures is left as a future work. We want to include more known results in the novelty filter to get this number of conjectures to an even more manageable number.

The main open task---and our current work---is closing the loop. The decisive gap is
proving: the formalization-and-proving stage is implemented and integrated
(\S\ref{sec:formalize-prove}) but has not yet been run on the candidate conjectures. Our immediate next
step is to run proof search over the surviving conjectures translated to Lean---together with a preamble-specific
lemma library to ground the provers on the custom invariants, which we anticipate to be the main obstacle
to closing the conjecture-proof loop because the provers we use were not trained on these invariants but on other parts of mathematics.

Mathematicians frequently also engage in hypothetical reasoning, they assume a conjecture is true and explore its consequences. For this reason, we also want to experiment with a \emph{Devil's Advocate} filter that would assume true the conjectures that survived the generate--refute and autoformalization and proving loops, and attempt to add them to knowledge base. We have portrayed this with semi-colon arrow in the pipeline diagram in Figure~\ref{fig:pipeline} but have not yet implemented it. 

Additionally, we want to explore the use of large language models as a \emph{conjecture critic} that would evaluate the plausibility of a candidate conjecture and provide feedback to the generator. This could help in prioritizing which conjectures to pursue further, especially when dealing with a large number of candidates. In fact, LLMs might be beneficial in the whole pipeline, not only as a critic but also as a generators and provers. \cite{davila2026graffiti3} has recently explored the use of LLMs in proposing new invariants. We want to keep exploring the use of LLMs in all stages of the pipeline.

\begin{acknowledgments}
  I thank Samuel Varchol and Vladimír Jančár for valuable discussions on techniques of counterexample search whose theses inspired that part of the work. I also thank anonymous reviewers for helpful suggestions and comments. 

  This work was supported by the use of computational resources of the supercomputer PERUN, operated by the Supercomputing Centre at the Technical University of Košice (TUKE), Slovakia with the support of the European Union from the funds of the Recovery and Resilience Plan of the Slovak Republic within the framework of project No. 17I03-04-P03-00001, Development and design of a supercomputer for the National Supercomputing Center.

\end{acknowledgments}

\section*{Declaration on Generative AI}
During the preparation of this work, the author used a large language model
as a coding and experimentation assistant to implement the pipeline, run the
experiments, and help with the draft of the manuscript. All computational results were verified against authoritative data sources -- House of Graphs and exact recomputation of GraphCalc; the
author reviewed and edited all content and takes full responsibility for the
publication's content.

\section*{Code availability}
All code and its documentation, data, logs, and databases are available and being developed at \url{https://github.com/JanPastorek/AutoGraphForge}.

\bibliography{sample-ceur}

\begin{thebibliography}{45}
\expandafter\ifx\csname natexlab\endcsname\relax\def\natexlab#1{#1}\fi
\providecommand{\url}[1]{\texttt{#1}}
\providecommand{\href}[2]{#2}
\providecommand{\path}[1]{#1}
\providecommand{\DOIprefix}{doi:}
\providecommand{\ArXivprefix}{arXiv:}
\providecommand{\URLprefix}{URL: }
\providecommand{\Pubmedprefix}{pmid:}
\providecommand{\doi}[1]{\href{http://dx.doi.org/#1}{\path{#1}}}
\providecommand{\Pubmed}[1]{\href{pmid:#1}{\path{#1}}}
\providecommand{\bibinfo}[2]{#2}
\ifx\xfnm\relax \def\xfnm[#1]{\unskip,\space#1}\fi
%Type = Inproceedings
\bibitem[{Larson and Cleemput(2017)}]{larson2017automated}
\bibinfo{author}{C.~E. Larson}, \bibinfo{author}{N.~V. Cleemput},
\newblock \bibinfo{title}{Automated conjecturing {I}: {Fajtlowicz}'s
  {Dalmatian} heuristic revisited (extended abstract)},
\newblock in: \bibinfo{booktitle}{Proceedings of the Twenty-Sixth International
  Joint Conference on Artificial Intelligence (IJCAI-17)},
  \bibinfo{year}{2017}.
%Type = Incollection
\bibitem[{DeLaVina(2005)}]{delavina2005history}
\bibinfo{author}{E.~DeLaVina},
\newblock \bibinfo{title}{Some history of the development of {Graffiti}},
\newblock in: \bibinfo{booktitle}{Graphs and Discovery},
  volume~\bibinfo{volume}{69} of \textit{\bibinfo{series}{DIMACS: Series in
  Discrete Mathematics and Theoretical Computer Science}},
  \bibinfo{publisher}{American Mathematical Society}, \bibinfo{year}{2005}, pp.
  \bibinfo{pages}{81--118}.
%Type = Article
\bibitem[{Tao(2025)}]{tao2025machine}
\bibinfo{author}{T.~Tao},
\newblock \bibinfo{title}{Machine-assisted proof},
\newblock \bibinfo{journal}{Notices of the American Mathematical Society}
  \bibinfo{volume}{72} (\bibinfo{year}{2025}) \bibinfo{pages}{6--13}.
%Type = Book
\bibitem[{Koretsky(2023)}]{koretsky2023ai}
\bibinfo{editor}{S.~Koretsky} (Ed.), \bibinfo{title}{Artificial Intelligence to
  Assist Mathematical Reasoning: Proceedings of a Workshop},
  \bibinfo{publisher}{National Academies Press}, \bibinfo{address}{Washington,
  DC}, \bibinfo{year}{2023}. \DOIprefix\doi{10.17226/27241}.
%Type = Article
\bibitem[{Fajtlowicz(1988)}]{fajtlowicz1988conjectures}
\bibinfo{author}{S.~Fajtlowicz},
\newblock \bibinfo{title}{On conjectures of {Graffiti}},
\newblock \bibinfo{journal}{Discrete Mathematics} \bibinfo{volume}{72}
  (\bibinfo{year}{1988}) \bibinfo{pages}{113--118}.
%Type = Misc
\bibitem[{{Wikipedia contributors}(2026)}]{wiki2026graffiti}
\bibinfo{author}{{Wikipedia contributors}}, \bibinfo{title}{Graffiti (program)
  --- {Wikipedia}{,} the free encyclopedia},
  \bibinfo{howpublished}{\url{https://en.wikipedia.org/wiki/Graffiti_(program)}},
  \bibinfo{year}{2026}.
%Type = Article
\bibitem[{Larson(2002)}]{larson2002intelligent}
\bibinfo{author}{C.~E. Larson},
\newblock \bibinfo{title}{Intelligent machinery and mathematical discovery},
\newblock \bibinfo{journal}{Graph Theory Notes of New York}
  \bibinfo{volume}{XLII} (\bibinfo{year}{2002}) \bibinfo{pages}{8--17}.
%Type = Article
\bibitem[{DeLaVina(2002)}]{delavina2002graffitipc}
\bibinfo{author}{E.~DeLaVina},
\newblock \bibinfo{title}{{Graffiti.pc}: A variant of {Graffiti}},
\newblock \bibinfo{journal}{Graph Theory Notes of New York}
  \bibinfo{volume}{XLII} (\bibinfo{year}{2002}) \bibinfo{pages}{26--30}.
%Type = Article
\bibitem[{Aouchiche et~al.(2009)Aouchiche, Hansen, and
  Stevanovi{\'c}}]{aouchiche2009variable}
\bibinfo{author}{M.~Aouchiche}, \bibinfo{author}{P.~Hansen},
  \bibinfo{author}{D.~Stevanovi{\'c}},
\newblock \bibinfo{title}{Variable neighborhood search for extremal graphs. 17.
  further conjectures and results about the index},
\newblock \bibinfo{journal}{Discussiones Mathematicae Graph Theory}
  \bibinfo{volume}{29} (\bibinfo{year}{2009}) \bibinfo{pages}{15--37}.
%Type = Article
\bibitem[{Caporossi and Hansen(2004)}]{caporossi2004graphedron}
\bibinfo{author}{G.~Caporossi}, \bibinfo{author}{P.~Hansen},
\newblock \bibinfo{title}{Variable neighborhood search for extremal graphs.
  {V}. three ways to automate finding conjectures},
\newblock \bibinfo{journal}{Discrete Mathematics} \bibinfo{volume}{276}
  (\bibinfo{year}{2004}) \bibinfo{pages}{81--94}.
%Type = Article
\bibitem[{M\'elot(2008)}]{melot2008facet}
\bibinfo{author}{H.~M\'elot},
\newblock \bibinfo{title}{Facet defining inequalities among graph invariants:
  The system {GraPHedron}},
\newblock \bibinfo{journal}{Discrete Applied Mathematics} \bibinfo{volume}{156}
  (\bibinfo{year}{2008}) \bibinfo{pages}{1875--1891}.
%Type = Misc
\bibitem[{Davila(2024)}]{davila2024automated}
\bibinfo{author}{R.~Davila}, \bibinfo{title}{Automated conjecturing in
  mathematics with {TxGraffiti}}, \bibinfo{year}{2024}.
  \href{http://arxiv.org/abs/2409.19379}{{\tt arXiv:2409.19379}}.
%Type = Article
\bibitem[{Caro et~al.(2022)Caro, Davila, Henning, and
  Pepper}]{caro2022txgraffiti}
\bibinfo{author}{Y.~Caro}, \bibinfo{author}{R.~Davila}, \bibinfo{author}{M.~A.
  Henning}, \bibinfo{author}{R.~Pepper},
\newblock \bibinfo{title}{Conjectures of {TxGraffiti}: Independence,
  domination, and matchings},
\newblock \bibinfo{journal}{Australasian Journal of Combinatorics}
  \bibinfo{volume}{84} (\bibinfo{year}{2022}) \bibinfo{pages}{258--274}.
%Type = Misc
\bibitem[{Davila(2024)}]{davila2024optimist}
\bibinfo{author}{R.~Davila}, \bibinfo{title}{The {Optimist}: Towards fully
  automated graph theory research}, \bibinfo{year}{2024}.
  \href{http://arxiv.org/abs/2411.09158}{{\tt arXiv:2411.09158}}.
%Type = Article
\bibitem[{Larson and Cleemput(2016)}]{larson2016dalmatian}
\bibinfo{author}{C.~E. Larson}, \bibinfo{author}{N.~V. Cleemput},
\newblock \bibinfo{title}{Automated conjecturing {I}: {Fajtlowicz's}
  {Dalmatian} heuristic revisited},
\newblock \bibinfo{journal}{Artificial Intelligence} \bibinfo{volume}{231}
  (\bibinfo{year}{2016}) \bibinfo{pages}{17--38}.
%Type = Misc
\bibitem[{Davila(2026)}]{davila2026graffiti3}
\bibinfo{author}{R.~Davila}, \bibinfo{title}{{Graffiti}$^3$: Compact theory
  libraries for automated mathematical discovery},
  \bibinfo{howpublished}{Preprint}, \bibinfo{year}{2026}.
  \DOIprefix\doi{10.13140/RG.2.2.20757.18402}.
%Type = Misc
\bibitem[{Bonte et~al.(2026)Bonte, Devillez, Dusollier, and
  M{\'e}lot}]{bonte2026phoeg}
\bibinfo{author}{S.~Bonte}, \bibinfo{author}{G.~Devillez},
  \bibinfo{author}{V.~Dusollier}, \bibinfo{author}{H.~M{\'e}lot},
  \bibinfo{title}{{PHOEG}: an online tool for discovery and education in
  extremal graph theory}, \bibinfo{year}{2026}.
  \href{http://arxiv.org/abs/2603.27242}{{\tt arXiv:2603.27242}}.
%Type = Article
\bibitem[{Christophe et~al.(2008)Christophe, Dewez, Doignon, Fasbender,
  Gr{\'e}goire, Huygens, Labb{\'e}, Elloumi, M{\'e}lot, and
  Yaman}]{christophe2008linear}
\bibinfo{author}{J.~Christophe}, \bibinfo{author}{S.~Dewez},
  \bibinfo{author}{J.-P. Doignon}, \bibinfo{author}{G.~Fasbender},
  \bibinfo{author}{P.~Gr{\'e}goire}, \bibinfo{author}{D.~Huygens},
  \bibinfo{author}{M.~Labb{\'e}}, \bibinfo{author}{S.~Elloumi},
  \bibinfo{author}{H.~M{\'e}lot}, \bibinfo{author}{H.~Yaman},
\newblock \bibinfo{title}{Linear inequalities among graph invariants: Using
  {GraPHedron} to uncover optimal relationships},
\newblock \bibinfo{journal}{Networks} \bibinfo{volume}{52}
  (\bibinfo{year}{2008}) \bibinfo{pages}{287--298}.
%Type = Article
\bibitem[{Brewster et~al.(1995)Brewster, Dinneen, and
  Faber}]{brewster1995computational}
\bibinfo{author}{T.~L. Brewster}, \bibinfo{author}{M.~J. Dinneen},
  \bibinfo{author}{V.~Faber},
\newblock \bibinfo{title}{A computational attack on the conjectures of
  {Graffiti}: New counterexamples and proofs},
\newblock \bibinfo{journal}{Discrete Mathematics} \bibinfo{volume}{147}
  (\bibinfo{year}{1995}) \bibinfo{pages}{35--55}.
%Type = Misc
\bibitem[{Jooken(2025)}]{jooken2025computer}
\bibinfo{author}{J.~Jooken}, \bibinfo{title}{Computer-assisted graph theory: a
  survey}, \bibinfo{year}{2025}. \href{http://arxiv.org/abs/2508.20825}{{\tt
  arXiv:2508.20825}}.
%Type = Misc
\bibitem[{Wagner(2021)}]{wagner2021constructions}
\bibinfo{author}{A.~Z. Wagner}, \bibinfo{title}{Constructions in combinatorics
  via neural networks}, \bibinfo{year}{2021}.
  \href{http://arxiv.org/abs/2104.14516}{{\tt arXiv:2104.14516}}.
%Type = Inproceedings
\bibitem[{de~Moura et~al.(2015)de~Moura, Kong, Avigad, Van~Doorn, and von
  Raumer}]{moura2015lean}
\bibinfo{author}{L.~de~Moura}, \bibinfo{author}{S.~Kong},
  \bibinfo{author}{J.~Avigad}, \bibinfo{author}{F.~Van~Doorn},
  \bibinfo{author}{J.~von Raumer},
\newblock \bibinfo{title}{The {Lean} theorem prover (system description)},
\newblock in: \bibinfo{booktitle}{International Conference on Automated
  Deduction}, \bibinfo{organization}{Springer}, \bibinfo{year}{2015}, pp.
  \bibinfo{pages}{378--388}.
%Type = Inproceedings
\bibitem[{van Doorn et~al.(2020)van Doorn, Ebner, and
  Lewis}]{vandoorn2020maintaining}
\bibinfo{author}{F.~van Doorn}, \bibinfo{author}{G.~Ebner},
  \bibinfo{author}{R.~Y. Lewis},
\newblock \bibinfo{title}{Maintaining a library of formal mathematics},
\newblock in: \bibinfo{booktitle}{Intelligent Computer Mathematics (CICM
  2020)}, \bibinfo{year}{2020}. \href{http://arxiv.org/abs/2004.03673}{{\tt
  arXiv:2004.03673}}.
%Type = Inproceedings
\bibitem[{Wu et~al.(2022)Wu, Jiang, Li, Rabe, Staats, Jamnik, and
  Szegedy}]{wu2022autoformalization}
\bibinfo{author}{Y.~Wu}, \bibinfo{author}{A.~Q. Jiang},
  \bibinfo{author}{W.~Li}, \bibinfo{author}{M.~N. Rabe},
  \bibinfo{author}{C.~Staats}, \bibinfo{author}{M.~Jamnik},
  \bibinfo{author}{C.~Szegedy},
\newblock \bibinfo{title}{Autoformalization with large language models},
\newblock in: \bibinfo{booktitle}{Advances in Neural Information Processing
  Systems (NeurIPS)}, \bibinfo{year}{2022}. \URLprefix
  \url{https://arxiv.org/abs/2205.12615}.
%Type = Misc
\bibitem[{Mavani and Pflueger(2026)}]{mavani2026chipfiring}
\bibinfo{author}{D.~D. Mavani}, \bibinfo{author}{N.~Pflueger},
  \bibinfo{title}{Formalizing chip-firing and {Riemann-Roch} for graphs in
  {Lean 4}}, \bibinfo{year}{2026}. \href{http://arxiv.org/abs/2606.16679}{{\tt
  arXiv:2606.16679}}.
%Type = Misc
\bibitem[{Nader et~al.(2026)Nader, Aljabea, Diehl, and
  Gupta}]{nader2026gtbench}
\bibinfo{author}{N.~Nader}, \bibinfo{author}{I.~Aljabea},
  \bibinfo{author}{P.~Diehl}, \bibinfo{author}{D.~Gupta},
  \bibinfo{title}{{GTBench}: A curriculum-grounded benchmark for evaluating
  {LLMs} as mathematical research assistants in graph theory},
  \bibinfo{year}{2026}. \href{http://arxiv.org/abs/2606.03144}{{\tt
  arXiv:2606.03144}}.
%Type = Misc
\bibitem[{Kalfus and Lidick{\'y}(2026)}]{kalfus2026ramsey}
\bibinfo{author}{J.~Kalfus}, \bibinfo{author}{B.~Lidick{\'y}},
  \bibinfo{title}{An automated proof that ${R(B_8, B_{10}) = 37}$},
  \bibinfo{year}{2026}. \href{http://arxiv.org/abs/2606.05629}{{\tt
  arXiv:2606.05629}}.
%Type = Misc
\bibitem[{Zhang et~al.(2026)Zhang, Sun, Suzuki, Lee, and
  Liu}]{zhang2026leanmarathon}
\bibinfo{author}{Y.~Zhang}, \bibinfo{author}{Y.~Sun},
  \bibinfo{author}{T.~Suzuki}, \bibinfo{author}{J.~D. Lee},
  \bibinfo{author}{F.~Liu}, \bibinfo{title}{{LeanMarathon}: Toward reliable
  {AI} co-mathematicians through long-horizon {Lean} autoformalization},
  \bibinfo{year}{2026}. \href{http://arxiv.org/abs/2606.05400}{{\tt
  arXiv:2606.05400}}.
%Type = Article
\bibitem[{{DeepSeek-AI}(2025)}]{deepseek2025prover}
\bibinfo{author}{{DeepSeek-AI}},
\newblock \bibinfo{title}{{DeepSeek-Prover-V2}: Scalable {MoE} transformer for
  formal theorem proving},
\newblock \bibinfo{journal}{DeepSeek Technical Report}  (\bibinfo{year}{2025}).
%Type = Article
\bibitem[{Coolsaet et~al.(2023)Coolsaet, D'hondt, and
  Goedgebeur}]{coolsaet2023house}
\bibinfo{author}{K.~Coolsaet}, \bibinfo{author}{S.~D'hondt},
  \bibinfo{author}{J.~Goedgebeur},
\newblock \bibinfo{title}{House of graphs 2.0: a database of interesting graphs
  and more},
\newblock \bibinfo{journal}{Discrete Applied Mathematics} \bibinfo{volume}{325}
  (\bibinfo{year}{2023}) \bibinfo{pages}{97--107}.
  \DOIprefix\doi{10.1016/j.dam.2022.10.013}.
%Type = Misc
\bibitem[{{M-A-P}(2026)}]{oprover2026}
\bibinfo{author}{{M-A-P}}, \bibinfo{title}{{OProver}: A unified framework for
  agentic formal theorem proving}, \bibinfo{year}{2026}.
  \href{http://arxiv.org/abs/2605.17283}{{\tt arXiv:2605.17283}},
  \bibinfo{note}{model: \url{https://huggingface.co/m-a-p/OProver-32B}}.
%Type = Article
\bibitem[{Davila(2025)}]{davila2025graphcalc}
\bibinfo{author}{R.~Davila},
\newblock \bibinfo{title}{{GraphCalc}: A {Python} package for computing graph
  invariants in automated conjecturing systems},
\newblock \bibinfo{journal}{Journal of Open Source Software}
  \bibinfo{volume}{10} (\bibinfo{year}{2025}) \bibinfo{pages}{8383}.
  \DOIprefix\doi{10.21105/joss.08383}.
%Type = Article
\bibitem[{{AIM Minimum Rank -- Special Graphs Work
  Group}(2008)}]{aim2008zeroforcing}
\bibinfo{author}{{AIM Minimum Rank -- Special Graphs Work Group}},
\newblock \bibinfo{title}{Zero forcing sets and the minimum rank of graphs},
\newblock \bibinfo{journal}{Linear Algebra and its Applications}
  \bibinfo{volume}{428} (\bibinfo{year}{2008}) \bibinfo{pages}{1628--1648}.
%Type = Inproceedings
\bibitem[{Hagberg et~al.(2008)Hagberg, Schult, and Swart}]{hagberg2008networkx}
\bibinfo{author}{A.~A. Hagberg}, \bibinfo{author}{D.~A. Schult},
  \bibinfo{author}{P.~J. Swart},
\newblock \bibinfo{title}{Exploring network structure, dynamics, and function
  using {NetworkX}},
\newblock in: \bibinfo{booktitle}{Proceedings of the 7th Python in Science
  Conference (SciPy 2008)}, \bibinfo{year}{2008}, pp. \bibinfo{pages}{11--15}.
%Type = Article
\bibitem[{Whitney(1932)}]{whitney1932congruent}
\bibinfo{author}{H.~Whitney},
\newblock \bibinfo{title}{Congruent graphs and the connectivity of graphs},
\newblock \bibinfo{journal}{American Journal of Mathematics}
  \bibinfo{volume}{54} (\bibinfo{year}{1932}) \bibinfo{pages}{150--168}.
%Type = Article
\bibitem[{K\H{o}nig(1931)}]{konig1931graphen}
\bibinfo{author}{D.~K\H{o}nig},
\newblock \bibinfo{title}{Gr\'afok \'es m\'atrixok},
\newblock \bibinfo{journal}{Matematikai \'es Fizikai Lapok}
  \bibinfo{volume}{38} (\bibinfo{year}{1931}) \bibinfo{pages}{116--119}.
%Type = Article
\bibitem[{Gallai(1959)}]{gallai1959uber}
\bibinfo{author}{T.~Gallai},
\newblock \bibinfo{title}{\"uber extreme punkt- und kantenmengen},
\newblock \bibinfo{journal}{Ann.\ Univ.\ Sci.\ Budapest, E\"otv\"os Sect.\
  Math.} \bibinfo{volume}{2} (\bibinfo{year}{1959}) \bibinfo{pages}{133--138}.
%Type = Article
\bibitem[{Allan and Laskar(1978)}]{allan1978domination}
\bibinfo{author}{R.~B. Allan}, \bibinfo{author}{R.~Laskar},
\newblock \bibinfo{title}{On domination and independent domination numbers of a
  graph},
\newblock \bibinfo{journal}{Discrete Mathematics} \bibinfo{volume}{23}
  (\bibinfo{year}{1978}) \bibinfo{pages}{73--76}.
%Type = Inproceedings
\bibitem[{Goedgebeur(2022)}]{goedgebeur2022introduction}
\bibinfo{author}{J.~Goedgebeur},
\newblock \bibinfo{title}{An introduction to computational graph theory and
  generation algorithms},
\newblock in: \bibinfo{booktitle}{CEUR Workshop Proceedings (ITAT'22)},
  \bibinfo{year}{2022}.
%Type = Inproceedings
\bibitem[{de~Moura and Bj{\o}rner(2008)}]{demoura2008z3}
\bibinfo{author}{L.~de~Moura}, \bibinfo{author}{N.~Bj{\o}rner},
\newblock \bibinfo{title}{{Z3}: An efficient {SMT} solver},
\newblock in: \bibinfo{booktitle}{Tools and Algorithms for the Construction and
  Analysis of Systems (TACAS 2008)}, volume \bibinfo{volume}{4963} of
  \textit{\bibinfo{series}{Lecture Notes in Computer Science}},
  \bibinfo{year}{2008}, pp. \bibinfo{pages}{337--340}.
  \DOIprefix\doi{10.1007/978-3-540-78800-3_24}.
%Type = Misc
\bibitem[{Damnjanovi\'c et~al.(2026)Damnjanovi\'c, Milivojevi\'c,
  \DJ{}or\dj{}evi\'c, and Stevanovi\'c}]{damnjanovic2026rlgt}
\bibinfo{author}{I.~Damnjanovi\'c}, \bibinfo{author}{U.~Milivojevi\'c},
  \bibinfo{author}{I.~\DJ{}or\dj{}evi\'c}, \bibinfo{author}{D.~Stevanovi\'c},
  \bibinfo{title}{{RLGT}: A reinforcement learning framework for extremal graph
  theory}, \bibinfo{year}{2026}. \href{http://arxiv.org/abs/2602.17276}{{\tt
  arXiv:2602.17276}}.
%Type = Inproceedings
\bibitem[{Kwon et~al.(2023)Kwon, Li, Zhuang, Sheng, Zheng, Yu, Gonzalez, Zhang,
  and Stoica}]{kwon2023vllm}
\bibinfo{author}{W.~Kwon}, \bibinfo{author}{Z.~Li},
  \bibinfo{author}{S.~Zhuang}, \bibinfo{author}{Y.~Sheng},
  \bibinfo{author}{L.~Zheng}, \bibinfo{author}{C.~H. Yu},
  \bibinfo{author}{J.~Gonzalez}, \bibinfo{author}{H.~Zhang},
  \bibinfo{author}{I.~Stoica},
\newblock \bibinfo{title}{Efficient memory management for large language model
  serving with {PagedAttention}},
\newblock in: \bibinfo{booktitle}{Proceedings of the 29th ACM Symposium on
  Operating Systems Principles (SOSP 2023)}, \bibinfo{year}{2023}, pp.
  \bibinfo{pages}{611--626}. \DOIprefix\doi{10.1145/3600006.3613165}.
%Type = Article
\bibitem[{Erd\H{o}s et~al.(1986)Erd\H{o}s, Saks, and S\'os}]{erdos1986induced}
\bibinfo{author}{P.~Erd\H{o}s}, \bibinfo{author}{M.~Saks},
  \bibinfo{author}{V.~T. S\'os},
\newblock \bibinfo{title}{Maximum induced trees in graphs},
\newblock \bibinfo{journal}{Journal of Combinatorial Theory, Series B}
  \bibinfo{volume}{41} (\bibinfo{year}{1986}) \bibinfo{pages}{61--79}.
%Type = Phdthesis
\bibitem[{Pepper(2004)}]{pepper2004thesis}
\bibinfo{author}{R.~Pepper}, \bibinfo{title}{Binding Independence}, Ph.D.
  thesis, University of Houston, \bibinfo{year}{2004}.
%Type = Article
\bibitem[{Pepper(2009)}]{pepper2009}
\bibinfo{author}{R.~Pepper},
\newblock \bibinfo{title}{On the annihilation number of a graph},
\newblock \bibinfo{journal}{Recent Advances in Electrical Engineering: Proc.\
  15th American Conf.\ on Applied Mathematics}  (\bibinfo{year}{2009})
  \bibinfo{pages}{217--220}.

\end{thebibliography}

\appendix

\section{Computational setup}
\label{app:compute}
\label{app:partitions}

\paragraph{Cluster hardware and budget.}
CPU workloads were deployed on HPE Cray XD2000 nodes ($2\times$ AMD EPYC 9745, $256$ cores,
$1{,}536$\,GB RAM, 200\,Gb/s NDR200 InfiniBand). The generate--refute experiment was a SLURM array of five independent
partitions, each on a full $256$-core node, run to a fixed point or a $29$\,h wall-clock cap; no partition
hit the cap---all five terminated at a fixed point. Per-round wall-time in generate-refute loop was
$\sim\!20$--$40$\,min, so a five-round partition finished in $\approx\!2$\,h and the
nine-round partition in $\approx\!4.5$\,h. The generate--refute loop itself therefore
accounts for only $256\times(4\times2+4.5)\approx3{,}200$ core-hours ($\approx\!0.37$
CPU-years). The remaining $\approx\!0.85$ of the $1.22$ CPU-year total is a one-time
cost dominated by the precomputation of  $59$ invariants with \textsc{GraphCalc} package
 over the $348{,}207$-graph refutation dataset---in particular the
exact-recompute (\texttt{bigdb}) tier, whose NP-hard ILP invariants (the
independence, domination, and zero-forcing families) dominate the per-graph
cost---together with the single-pass baseline run of \S\ref{sec:dynamics}. The
total computational cost was thus $1.22$ CPU-years ($\approx\!0.37$ generate--refute,
$\approx\!0.85$ battery precomputation and baseline).

Since the precomputation dominates, and within it the NP-hard invariants
dominate, replacing the ILP formulations with a SAT or CP-SAT encoding is the
single most promising engineering change available to us. Independence,
domination and zero-forcing all admit natural propositional encodings, and
modern conflict-driven and local-search solvers are frequently far faster than
ILP on instances of this size. A cheaper battery is not merely a saving: it
would let us extend the refutation tiers to larger orders, and so reach
invariants that are uninformative on the small graphs the present pool is
concentrated on. We are grateful to a reviewer for this suggestion.

\paragraph{Target-invariant partitioning.}
The $45$ numeric target invariants were sorted alphabetically by their
\textsc{GraphCalc} identifier and sliced into five contiguous blocks of nine; partition~$i$
(Table~\ref{tab:rounds}) generated bounds with the invariants of block~$i$ on the left-hand
side. This split is a deterministic load-balancing device, not a semantic grouping---alphabetical
ordering deliberately scatters related families (the domination variants, for instance, fall across
partitions~1--3)---so that the five full-node jobs carry comparable work and the assignment is
reproducible from the public target list. The complete assignment is listed below.

\begin{description}
  \item[Partition 0] algebraic connectivity, annihilation number, average degree, average
    shortest path length, burning number, chromatic number ($\chi$), clique number ($\omega$),
    connected zero-forcing number, diameter.
  \item[Partition 1] domination number ($\gamma$), double Roman domination number, edge-cover
    number, harmonic index, independence number ($\alpha$), independent domination number ($i$),
    largest Laplacian eigenvalue, matching number ($\nu$), maximum degree ($\Delta$).
  \item[Partition 2] minimum maximal matching number, minimum degree ($\delta$), order ($n$),
    outer-connected domination number, positive-semidefinite zero-forcing number, power domination
    number, radius, residue, restrained domination number.
  \item[Partition 3] Roman domination number, second-largest adjacency eigenvalue, size ($m$),
    Slater number, smallest adjacency eigenvalue, spectral radius ($\rho$), sub-total domination
    number, three-rainbow domination number, total domination number.
  \item[Partition 4] total zero-forcing number, triameter, two-forcing number, two-rainbow
    domination number, vertex clique-cover number, vertex-cover number ($\tau$), well-splitting
    number, zero-adjacency-eigenvalue count, zero-forcing number.
\end{description}

\section{Counterexample-search backend settings}
\label{app:hyperparams}
The hyperparameters of the active-search backends (\S\ref{sec:backends}) were not optimised and were used unchanged across all
conjectures. All backends are seeded from the structured base library (\S\ref{sec:data}) and run
under a per-candidate wall-clock cap.

\begin{table}[htbp]
  \caption{Backend hyperparameters (standard defaults, not tuned per conjecture).}
  \label{tab:hyperparams}
  \begin{tabular}{ll}
    \toprule
    Backend & Settings \\
    \midrule
    \textsf{cross\_entropy} & per-edge probability matrix, batch sampling; elite momentum $0.3$ \\
    \textsf{vns}            & $k$-edge shake + greedy $1$-edge ascent up to $6$ steps; reset $k{=}1$ on improvement \\
    \textsf{mcts}           & UCT bandit, $c=\sqrt{2}\approx1.41$; random $1$-edge rollouts to depth $6$ \\
    \textsf{sa}             & $80\%$ $1$-edge / $20\%$ $2$-edge flips; linear cooling $T_0=1.0\to T_{\min}=10^{-3}$ \\
    \textsf{rlgt-RL}        & Deep Cross-Entropy policy; reward $=$ violation margin \\
    \bottomrule
  \end{tabular}
\end{table}

In the HPC run the search was swept over a wide order set from $n = 10$ to $n = 100$, with the per-order number of trials scaled down as the order grows but floored so that even the largest orders receive a meaningful number of attempts, and with a per-candidate wall-clock budget that bounds round time regardless of how expensive the exact invariants become at
large $n$; this lets the engine search for counterexamples on bigger orders while
keeping each round time-bounded.

\end{document}